\documentclass[10pt,twocolumn,letterpaper]{article}

\usepackage{xcolor}
\definecolor{tsinghuaPurple}{RGB}{77,25,121}

\usepackage{iccv}              

\usepackage{multirow}

\definecolor{iccvblue}{rgb}{0.21,0.49,0.74}
\usepackage[pagebackref,breaklinks,colorlinks,allcolors=iccvblue]{hyperref}

\usepackage{bibunits}

\def\paperID{14093} 
\def\confName{ICCV}
\def\confYear{2025}

\title{INSTINCT: Instance-Level Interaction Architecture for Query-Based Collaborative Perception}

\author{Yunjiang Xu$^{1}$, Lingzhi Li$^{1}$\thanks{Corresponding author.}, Jin Wang$^{2}$\footnotemark[1], Yupeng Ouyang$^{1}$, Benyuan Yang$^{2}$\\
$^{1}$School of Computer Science and Technology, Soochow University\\
$^{2}$School of Future Science and Engineering, Soochow University\\
{\tt\small \{yjxu09, ouyangchina1\}@gmail.com, \{lilingzhi, wjin1985, byyang\}@suda.edu.cn}
}

\begin{document}
\maketitle
\begin{abstract}
Collaborative perception systems overcome single-vehicle limitations in long-range detection and occlusion scenarios by integrating multi-agent sensory data, improving accuracy and safety. 
However, frequent cooperative interactions and real-time requirements impose stringent bandwidth constraints. 
Previous works proves that query-based instance-level interaction reduces bandwidth demands and manual priors, however, LiDAR-focused implementations in collaborative perception remain underdeveloped, with performance still trailing state-of-the-art approaches. 
To bridge this gap, we propose \textbf{INSTINCT} (\textbf{INST}ance-level \textbf{IN}tera\textbf{C}tion Archi\textbf{T}ecture), a novel collaborative perception framework featuring three core components: 
1) a quality-aware filtering mechanism for high-quality instance feature selection; 
2) a dual-branch detection routing scheme to decouple collaboration-irrelevant and collaboration-relevant instances; 
and 3) a Cross Agent Local Instance Fusion module to aggregate local hybrid instance features. 
Additionally, we enhance the ground truth (GT) sampling technique to facilitate training with diverse hybrid instance features. 
Extensive experiments across multiple datasets demonstrate that INSTINCT achieves superior performance.
Specifically, our method achieves an improvement in accuracy 13.23\%/33.08\% in DAIR-V2X and V2V4Real while reducing the communication bandwidth to 1/281 and 1/264 compared to state-of-the-art methods. 
The code is available at \textcolor{tsinghuaPurple}{https://github.com/CrazyShout/INSTINCT.}
\end{abstract}    
\section{Introduction}
\label{sec:intro}

\begin{figure}[htbp]
  \centering
   \includegraphics[width=\columnwidth, trim=18pt 15pt 15pt 15pt, clip]{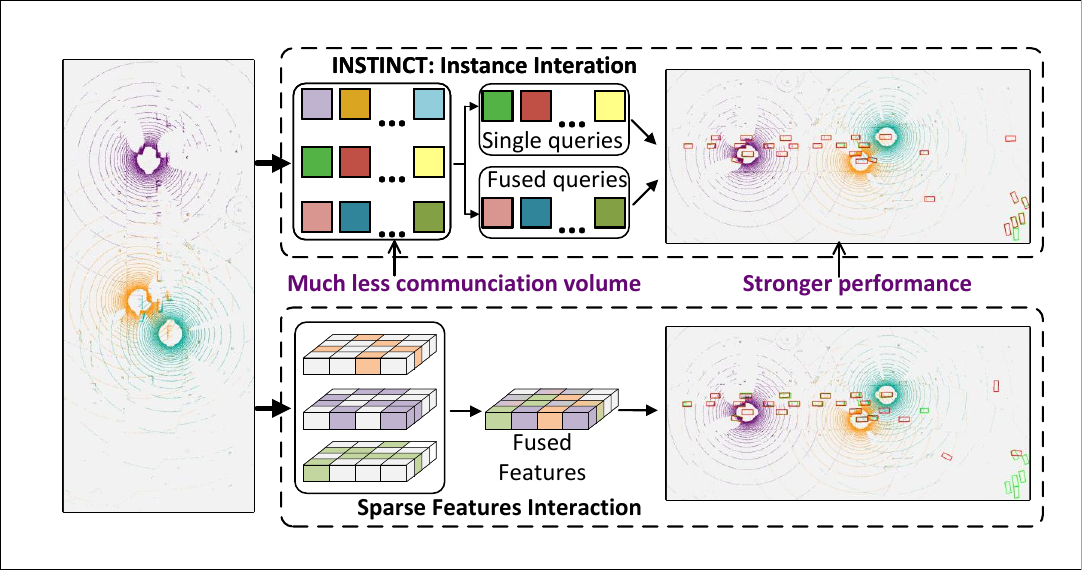}

   \caption{
   By implementing the innovative instance interaction architecture, INSTINCT demonstrates superior performance while operating under lower bandwidth constraints.
   }
   \label{fig:onecol}
\end{figure}


Perception is critical in autonomous driving, providing accurate inputs for decision-making and control. LiDAR has emerged as a key sensor for object detection, thanks to its exceptional 3D perception capabilities \cite{lang2019pointpillars,wu2023transformation,wu2022sparse}. However, traditional single-agent perception systems are constrained by limited sensing range and inherent sensor deficiencies, leading to blind spots in obstructed scenarios and adverse weather conditions, posing significant safety risks.

Multi-view and multi-agent collaborative perception addresses these limitations by utilizing distributed sensor networks. For example, infrastructure-based sensors, such as cameras and LiDAR, typically offer higher resolution, improved accuracy, and wider coverage, enabling effective detection of distant objects. Currently, collaborative perception methods are classified into early fusion, intermediate fusion, and late fusion, with intermediate fusion showing the greatest potential. Notably, extensive research has focused on Vehicle-to-Everything (V2X) systems \cite{liu2020when2com,xu2022cobevt,xu2022v2x}, integrating vehicle and infrastructure data to significantly enhance perception performance.


However, as shown in Fig.\ref{fig:onecol}, the requirement for real-time perception ($\ge$10Hz) poses a severe challenge to communication bandwidth. Traditional intermediate fusion methods directly transmit complete feature maps \cite{wang2020v2vnet,xu2022v2x,mehr2019disconet}, resulting in enormous bandwidth pressure. 
Where2comm \cite{hu2022where2comm} filters key information by constructing request and requirement graphs, and CodeFilling \cite{hu2024communication} further introduces a learning-based codebook compression mechanism, but the transmission of sparse feature maps still incurs high communication costs.

It is worth noting that query-based methods have been widely studied in single agent perception, and using query features for interaction in collaborative perception is also known as instance level fusion \cite{fan2024quest,chen2023transiff,wang2024iftr,huang2024actformer}.
It is obviously that sending instances features only is accurate and bandwidth friendly. 
However, current instance interaction methods are mainly focused on camera modality, and the unique representation of LiDAR point clouds limit their direct transfer.
Although TransIFF \cite{chen2023transiff} has achieved instance level collaborative perception of LiDAR for the first time, its architecture only supports vehicle-infrastructure (V2I) collaborative scenarios. 
In addition, TransIFF emphasizes communication bandwidth, but its performance still lags significantly behind advanced models.

To eliminate the problems we mentioned, we proposed INSTINCT, a novel \textbf{INST}ance-level \textbf{IN}tera\textbf{C}tion Archi\textbf{T}ecture based on LiDAR-V2X Systems.
This architecture achieves breakthroughs through three innovative design:
1)
An instance feature filtering mechanism based on selective transmission, effectively reducing communication bandwidth;
2)
The dual branch detection architecture achieves collaborative/single agent detection decoupling, eliminating irrelevant instance interference;
3)
The Cross-Agent Local Instance Fusion (CALIF) module achieves feature optimization of hybrid instances through domain adaption and Gaussian-distance-based local fusion.

We evaluate INSTINCT on multiple datasets, such as DAIR-V2X \cite{yu2022dair}, V2XSet \cite{xu2022v2x}, and V2V4Real \cite{xu2023v2v4real}.
It shows that INSTINCT achieves state-of-the-art performance in all metrics while maintaining a low communication load of $2^{13}-2^{14}$ bytes/frame, especially on the DAIR-V2X and V2V4Real real-world datasets.
The main contributions can be summarized as follows:
\begin{itemize}[leftmargin=\leftmarginii]
    \item We propose INSTINCT, the first collaborative perception architecture enabling LiDAR-based instance-level interaction in V2X scenarios, achieving a superior balance between bandwidth usage and perception performance.
    \item We design a filtering mechanism to efficiently communicate instance information, combined with a dual-branch detection strategy for single-agent and collaborative detection, effectively avoiding irrelevant instance interference and improving system robustness.
    \item We introduce a Cross-Agent Local Instance Fusion module, effectively addressing domain gaps and leveraging local correlations among instances. INSTINCT achieves state-of-the-art performance across multiple datasets while maintaining a minimal bandwidth usage of only $2^{13}$-$2^{14}$ bytes.
\end{itemize}


\begin{figure*}
  \centering
  \includegraphics[width=\textwidth, trim=6.6cm 0.86cm 8.2cm 0.6cm, clip]{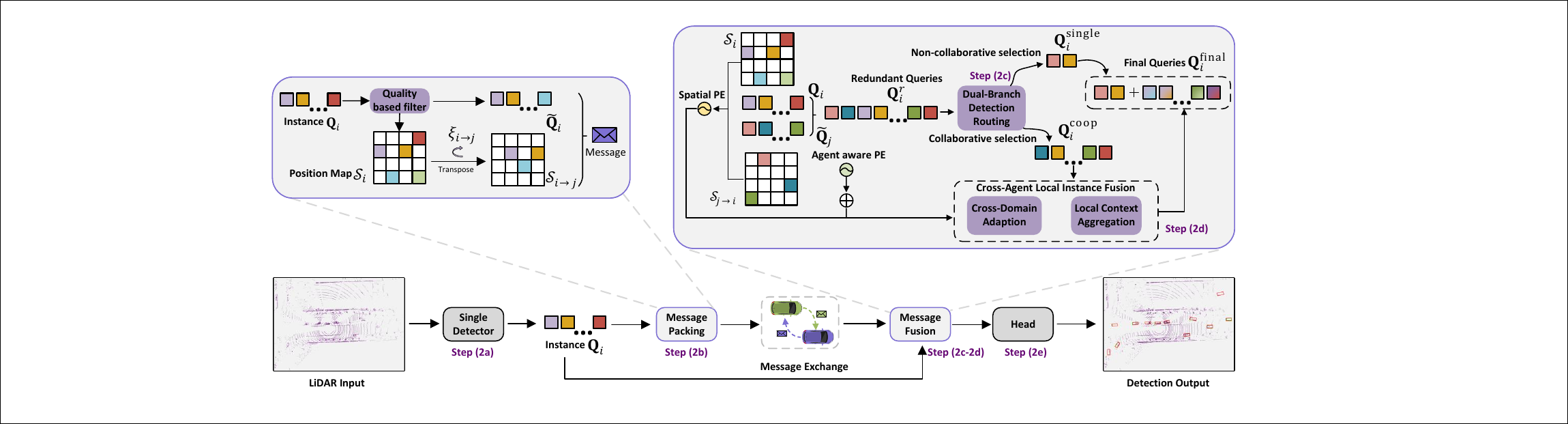}
  \caption{\textbf{System Overview.}
  Given LiDAR data for each intelligent agent, INSTINCT first extracts instance-level features using a single-agent detector. 
  Next, it filters these features and attaches a spatial position map before transmitting them to other agents.
  Upon receiving messages from other agents, INSTINCT integrates its own features with the incoming features.
}
  \label{overview}
    
\end{figure*}


\section{Related Work}

\subsection{Collaborative Perception}
Collaborative perception leverages multi-agent interactions to enhance individual vehicle sensing capabilities and driving safety. The availability of high-quality datasets like DAIR-V2X \cite{yu2022dair}, V2XSet \cite{xu2022v2x}, V2V4Real \cite{xu2023v2v4real}, and OPV2V \cite{xu2022opv2v} has significantly advanced related research.
Some studies focus on reducing communication bandwidth while maintaining performance. Where2comm \cite{hu2022where2comm} employs spatial confidence maps to eliminate irrelevant information, concentrating on perceptually critical areas. CodeFilling \cite{hu2024communication} further compresses data via a codebook-based representation, selectively transmitting essential information.
Pose errors negatively impact perception performance. CoAlign \cite{lu2023robust} addresses this through pose graph modeling, while FreeAlign \cite{lei2024robust} integrates graph neural networks and multi-anchor subgraph search to enhance robustness against pose inaccuracies.
To address spatiotemporal misalignment from communication latency, FFNet \cite{yu2024flow} introduces feature flow prediction for temporal alignment. CoBEVFlow and CoDynTrust \cite{wei2024asynchrony, xu2025codyntrust} propose BEV-flow-based motion compensation to align ROI features, ensuring effective spatiotemporal synchronization.

\subsection{Query-based Object Detection}
DETR \cite{carion2020end} pioneered an end-to-end detection paradigm by reconstructing the detection process with an attention mechanism and a query stream, significantly simplifying the complex procedures of traditional object detection. 
Subsequent studies have focused on its optimization.
For instance, Deformable-DETR \cite{zhu2020deformable} achieves efficient local interactions through a deformable attention mechanism, dramatically reducing computation, and DN-DETR \cite{li2022dn} introduces a denoising training strategy to accelerate model convergence. 
Building on the success of DETR-based method in 2D detection, its ideas have been extended to the 3D detection domain.
Specifically, for camera modalities, BEVFormer \cite{li2024bevformer} constructs a grid-based query structure in a bird’s-eye view to enhance 3D spatial encoding.
SparseBEV \cite{wu2022sparse} introduces scale-adaptive attention along with a hybrid spatiotemporal sampling strategy to achieve fully sparse BEV detection. 
For LiDAR modalities, ConQueR \cite{zhu2023conquer} pioneered a query contrast strategy to eliminate dense false positives, and SEED \cite{liu2024seed} enhances detection accuracy by reconstructing the query selection and interaction modules.
\\


\section{Problem formulation}
Consider $N$ agents in the scene, where each agent is able to send and receive information to and from other agents. Let $\mathcal{X}_i$ be the raw input data for $i$-th agent, and $\mathcal{Y}_i$ is the corresponding ground truth in the scene.
The objective is to maximize the detection performance of all agents with total communication cost $B$; that is,
\begin{equation}
    max\sum_i^N{g(f_\theta(\mathcal{X}_i,\{\mathcal{P}_{j\to i}\}_{j=1}^N),\mathcal{Y}_i)} \quad \textit{s.t.} \sum_{i=1}^{N}\lvert \mathcal{P}_{i\to j}\rvert \leq B,
\end{equation}
where $g(\cdot,\cdot)$ is the detection evaluation metric, $f$ is the collaborative perception network with trainable parameter $\theta$, and $\mathcal{P}_{j\to i}$ denotes the message transmitted from $j$-th agent to $i$-th agent.

\section{Methodology}

\subsection{Overall}
For the $i$-th agent, the proposed collaborative works as follows:
\begin{subequations}
\begin{flalign}
&\mathbf{Q}_{i} = f_{\text{single}}(\mathcal{X}_i)                             \label{query} \\
&\tilde{\mathbf{Q}}_i, \mathcal{S}_{j \rightarrow i} = f_{\text{filter}}(\mathbf{Q}_i, \mathcal{\xi}_{j\rightarrow i })                \label{filter} \\
&\mathbf{Q}_i^{\text{single}}, \mathbf{Q}_i^{\text{coop}} = f_{\text{DDR}}(\mathbf{Q}_i, \tilde{\mathbf{Q}}_j) \label{DDR_eq} \\
&\mathbf{Q}_i^{\text{final}} = f_{\text{CALIF}}(\mathbf{Q}_i^{\text{coop}},\mathcal{S}_i,\mathcal{S}_{j \rightarrow i}) + \mathbf{Q}_i^{\text{single}} \label{final} \\
&\hat{\mathcal{Y}}_i = f_{\text{head}}(\mathbf{Q}_i^{\text{final}}) \label{head}
\end{flalign}
\end{subequations}

As shown in Fig.\ref{overview}, in step \eqref{query}, $\mathbf{Q}_i \in \mathbb{R}^{c}$ denotes queries extracted from $\mathcal{X}_i$ by $f_\text{single}$.
In step \eqref{filter}, $f_\text{filter}$ means Filtering Module that filter $\mathbf{Q}_i$.  $\mathbf{\tilde{Q}}_i$ denotes filtered queries features and $S_{j\to i}$ denotes the spatial position map $j$-th agent coordinate to $i$-th agent, which will be discussed in \ref{qulity filtering}.
$\xi_{j\rightarrow i} \in \mathbb{R}^{4\times4}$ denotes spatial transform matrix.

After receiving $\mathbf{\tilde{Q}}_j$ sent from $j$-th agent, $i$-th agent start to collaborate.
In step \eqref{DDR_eq}, $i$-th agent send $\mathbf{\tilde{Q}_j}$ and $\textbf{Q}_i$ into the DDR (Dual-Branch Detection Routing) module, separately computing the collaborative queries and non-collaborative queries. 
Next, at step \eqref{final}, CALIF (Cross-Agent Local Instance Fusion module) aggregates collaborative features, performs cross-domain adaptation adjustments, and finally conducts local instance interactions. 
Then, at step \eqref{head}, all concatenated instance-level features $\mathbf{Q}_i^\text{final}$ are passed through a detection head to obtain the final detection result $\hat{\mathcal{Y}}_i$.

\subsection{Single-Agent Detector}\label{single}
As the foundation of collaborative perception, single-agent detection provides essential perceptual information for the collaboration process.
Inspired by ConQueR \cite{zhu2023conquer}, we use BoxAttention \cite{nguyen2022boxer} as a key component of both the Encoder and Decoder, where the Encoder is responsible for feature extraction and the Decoder corrects the reference boxes.

To better supervise training, we also modify the loss function for supervising the final output, which will be explained in Section \ref{qulity filtering}.
Additionally, a comparison with advanced single-vehicle models for direct post-fusion is listed in the appendix.
In summary, following the single-agent detection process, each agent produces unfiltered outputs.
\subsection{Quality-Aware Filtering}\label{qulity filtering}
After the single-agent detection stage, the collaboration process determines the optimal moment for cooperation based on predefined rules, such as a simple distance-triggered condition.
Specifically, when the distance between two agents falls below a preset threshold, collaboration is triggered, executing steps \eqref{filter} to \eqref{head}.

In step \eqref{filter}, we perform the following filtering operation:
1) the preparation of high-quality collaborative instance features to avoid unnecessary interference during the collaboration; 
and 2) the preparation of a unified spatial position marker for each instance feature to ensure that, in subsequent steps, the system can understand the relative positions between all instances.
Additionally, based on the unified spatial position markers, instance features outside the ego vehicle's perception range can be excluded from the communication process in advance.
Both of these operations help filter unnecessary features, which further reduces the transmission bandwidth.

Recent studies \cite{zhao2024detrs, liu2024seed} have made corresponding improvements in query selection.
Since confidence scores only reflect the credibility of classification but do not capture the reliability of box regression, it can lead to an unfair query selection process.
To ensure that the quality of the instance features transmitted during information exchange is sufficiently high, we consider two methods: 1) predicting the IoU score with a separate branch, and 2) applying an IoU-related penalty to the classification loss.
Ultimately, we chose the second approach.
This decision was based on the different performance of the separate branch IoU prediction across both synthetic and real-world datasets, with related analysis presented in the appendix.
Specifically, we apply an IoU-based penalty using the MAL \cite{huang2024deim} loss in the final layer of the single-object decoder.
The loss formula is as follows:
\begin{equation}
\text{MAL}(p, q, y) = \begin{cases} 
-q^\gamma \log(p) + (1 - q^\gamma) \log(1 - p) & q > 0 \\
-p^\gamma \log(1 - p) &  q = 0 
\end{cases}
\end{equation}

Here, \( p \) and \( q \), ranging within $[0, 1]$, denote the foreground prediction probability and the IoU between the predicted and ground-truth boxes, respectively. The tunable parameter \( \gamma \) balances the gradient contributions from hard and easy samples. Notably, while the MAL loss was originally designed for 2D detection, its adaptation to 3D point cloud detection introduces heightened numerical sensitivity due to the precise alignment of eight corner points in 3D IoU (compared to four vertices in 2D IoU), often causing training instability. This issue is particularly pronounced when MAL is applied across multiple decoder layers. To address this, we apply MAL solely in the final layer and employ the computationally stable Bird’s Eye View (BEV) IoU as the baseline for \( q \).

To reduce the amount of communication bandwidth, it is a natural idea to only transmit the perception instances that inside the ego's perception range, which aligns with the idea of having a unified location marker based on a coordinate system.
Therefore, we first label the position of each instance feature in its spatial domain, creating a sparse 2D relative position map $\mathcal{S}_i$.
Then, we apply a spatial transformation matrix $\xi_{i\rightarrow j}$ to transform this position map.
Based on the transformed position map, we obtain $\tilde{\mathbf{Q}}_i$ and the corresponding unified spatial position map $\mathcal{S}_{i\rightarrow j}$.
Afterward, $\tilde{\mathbf{Q}}_i$ and $\mathcal{S}_{i\rightarrow j}$ are packaged and transmitted to other agents.

Overall, through QAF (Quality-Aware Filtering) module, we effectively select features that need to transmit, aligned coordinates spatially, reducing the transmission bandwidth and improving accuracy.

\subsection{Dual-branch Detection Routing}\label{CDA}
It is important to note that step \eqref{DDR_eq} is based on the intuition that if an instance does not have any corresponding instances in the entire collaborative scene, it cannot gain any benefit from the collaboration process. 
Conversely, if this instance is not distinguished and is directly placed into the collaboration, it may negatively affect the collaboration. 
To address this, the following steps are executed:
\begin{enumerate}
    \item First, all received instance features are concatenated into an instance vector table $\mathbf{Q}_i^r$, which may contain redundant element.
    \item Next, $\mathbf{Q}_i^r$ is sent to a detection head, obtaining detection results. 
    Here, the Head shares parameters from the final output layer of the single-agent detector.
    \item Finally, we compute the IoU between all pairs of vectors in the vector table, forming an IoU matrix $\mathcal{M}_{iou}$. 
    After removing the diagonal entries, we filter the rows of $\mathcal{M}_{iou}$ where all values are below a threshold $\mathcal{\lambda}$, and retrieve the corresponding instances from $\mathbf{Q}_i^{r}$ to form $\mathbf{Q}_i^{single}$. The remaining instances constitute $\mathbf{Q}_i^{coop}$, where $\mathcal{\lambda}$ is a hyperparameter threshold.
\end{enumerate}

Through the above steps, we divide all instance features into two categories: one as $\mathbf{Q}_i^{single}$ that do not require collaboration, and the other as $\mathbf{Q}_i^{coop}$ that are potentially need fusion for collaboration.

\begin{figure}[htbp]
  \centering
  \includegraphics[width=\columnwidth, trim=160pt 18pt 233pt 18pt, clip]{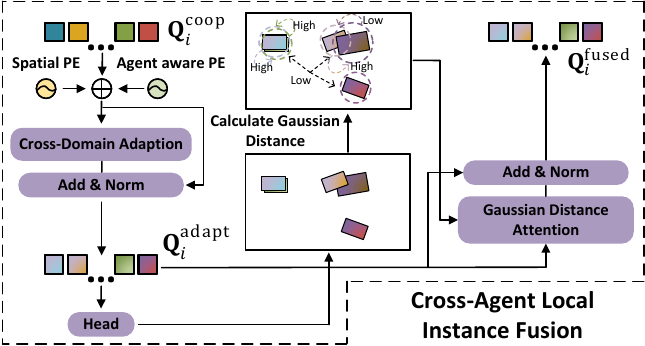}
  \caption{\textbf{Overall Structure of CALIF.}
  In this module, instance features are added with position encoding, then forwarded to the CDA module.
  Its results are used to compute Gaussian distances, identifying relevant features to obtain $\mathbf{Q}_i^{fused}$.
  }
  \label{CALIF}
\end{figure}

\begin{table*}[ht]
\centering
\begin{tabular*}{\linewidth}{c|c|ccc|c}
\hline
\multirow{2}{*}{Model} & \multirow{2}{*}{Fusion Type} & DAIR-V2X      & V2XSet     & V2V4Real      & {Comm(log2)} \\ \cline{3-5}
                       &                            & AP@0.5/0.7 ↑    & AP@0.5/0.7 ↑ & AP@0.5/0.7 ↑ &  DAIR-V2X/V2XSet/V2V4Real     \\ \hline
No Fusion              & \textbackslash{}           & 0.6349/0.4962 & 0.6515/0.5204  & 0.3980/0.2200  & 0                       \\ 
Late Fusion            & \textbackslash{}           & 0.6043/0.3746 & 0.7876/0.6801  & 0.5500/0.2670  &  8.39/9.60/9.79                      \\
V2VNet\cite{wang2020v2vnet}      & Intermidate                 & 0.6344/0.4227 & 0.8274/0.6582  & 0.6470/0.3360  & 24.62/25.10/25.10         \\
V2XViT\cite{xu2022v2x}           & Intermidate                 & 0.7349/0.5675 & 0.8431/0.7018  & 0.6490/0.3690  & 24.62/25.10/25.10         \\
DiscoNet\cite{mehr2019disconet}  & Intermidate                 & 0.7469/0.5920 & 0.8778/0.7207  & 0.6412/0.3451  & 24.62/25.10/25.10         \\
CoBEVT\cite{xu2022cobevt}&  Intermidate                & 0.6825/0.5787 & 0.8454/0.7119 & 0.5556/0.2708  & 24.62/25.10/25.10         \\
Where2comm\cite{hu2022where2comm}&  Intermidate                & 0.7901/0.6649 & 0.9259/0.8492 & 0.7021/0.3801  & 21.72/21.19/22.86         \\
CoAlign\cite{lu2023robust}                       & Intermidate                           & 0.7797/0.6547  & \textbf{0.9293}/0.8466  & 0.7209/0.4656  & 24.62/25.10/25.10                           \\
\textbf{Ours}          & Instance                           & \textbf{0.8191}/\textbf{0.7529}  & 0.9229/\textbf{0.8731}  & \textbf{0.8088}/\textbf{0.6196}  &  \textbf{13.58/14.16/14.81}        \\ \hline
\end{tabular*}
\caption{Different models' performance on the different datasets.}\label{tab_performace}
\end{table*}

\subsection{Cross-Agent Local Instance Fusion} \label{instance_fusion}
Step \eqref{final} addresses the fusion of collaborative mixed instance features $\mathbf{Q}_i^{coop}$, where two components are designed to resolve two practical challenges:
1)
To mitigate inevitable domain gaps arising from heterogeneous hardware configurations across agents and environmental disparities, we need to introduce a cross-domain interaction mechanism.
2)
Given the extreme sparsity of LiDAR data distributions, global interaction is computationally redundant and risks introducing extraneous learning tasks.
Thus, we need local instance interaction mechanism.

Furthermore, for any two instance features \(p\) and \(q\) in \(\mathbf{Q}_i^{coop}\), the mutual information exchange during agent collaboration exhibits inherent asymmetry. 
More specifically, the degree of interaction should depend on the differences between them in order to complement the completeness of each feature’s information. 
Therefore, the interactions among the mixed features must be local and asymmetric.

For the first challenge, a self-attention mechanism is employed to bridge domain gaps, formulated as:
\begin{equation}
    \begin{split}
        \mathbf{Q}_i^{adapt} &=\text{CDA}(\mathbf{Q}_i^{coop}, \mathbf{K}_i^{coop},\mathbf{V}_i^{coop}) \\
        &= \text{Softmax}(\frac{\mathbf{Q}_i^{coop} (\mathbf{K}_i^{coop})^T}{ \sqrt{d_k}}) \cdot \mathbf{V}_i^{coop},
    \end{split}
\end{equation}
where CDA denotes Cross-Domain Adaption, and $\mathbf{K}_i^{coop} = \mathbf{V}_i^{coop} = \mathbf{Q}_i^{coop}$. 
This operation adjusts the data distribution of each instance in the mixed samples, thereby bridging the domain gap. 
It is important to note that before cross-domain adaptation, dual position encodings were added to each instance.
One of the position encoding is the unified spatial position coordinate map from the cooperative information, which serves as the spatial position encoding. 
The other one is the agent perception encoding, where we assign the ego agent the identifier 0, and the identifiers for the remaining agents are determined sequentially based on the timing of the cooperative process. 
Finally, we encode the perception identifiers to obtain the agent perception position encoding. 
These two position encodings enable the interaction module to identify the spatial location of an instance and the corresponding agent it belongs to.

The above operation is not sufficient to optimize each instance; we also need to further calibrate the instances through stronger local relational interactions while masking irrelevant instances.
Inspired by TransFusion \cite{bai2022transfusion}, we designed a local attention mechanism based on Gaussian distance, as illustrated in Fig. \ref{CALIF}. 
This is a heuristic approach in which all mixed instances are through a shared parameter Head, obtaining the corresponding detection boxes for each instance. 
Then,a circumcircle for each instance's bounding box is generated. 
For a given instance feature, we focus only on the distance from the center $(x,y)$ of the bounding boxes of other instance features to the center of the circle. 
This distance guides to generate an attention weight, as shown in the following equation:
\begin{equation}
    \mathcal{W}_{k,v} = \exp(-\frac{\sqrt{(x_k - x_v)^2 + (y_k - y_v)^2}}{\beta r_k ^ 2}),
\end{equation}
where $(x_k,y_k)$ and $(x_v,y_v)$ denotes the centers of two corresponding detection boxes in the mixed instances.
Notice that $\mathcal{W}_{k,v} \in (0,1]$, and $\beta$ is a hyperparameter used to adjust the range of attention.

\begin{equation}
    \begin{split}
        \mathbf{Q}_i^{fused} =\ &\text{GDA}(\mathbf{Q}_i^{adapt},\mathbf{K}_i^{adapt},\mathbf{V}_i^{adapt}) \\
        =\ & \text{Softmax}(\frac{\mathbf{Q}_i^{adapt} (\mathbf{K}_i^{adapt})^T}{ \sqrt{d_k}}\\
        & + \log(\mathcal{W})) \cdot \mathbf{V}_i^{adapt},
    \end{split}
\end{equation}
where GDA denotes Gaussian Distance Attention, and $\mathbf{K}_i^{adapt} = \mathbf{V}_i^{adapt} = \mathbf{Q}_i^{adapt}.$

Through this attention mechanism, features will only focus on local features that are close to them, while those that are far away or have large prediction biases will be ignored.
In the end, we combine $\mathbf{Q}_i^{fused}$ and $\mathbf{Q}_i^{single}$ into a single entity, $\mathbf{Q}_i^{final}$.

\begin{figure*}[ht]
    \centering
    \includegraphics[width=1.0\linewidth]{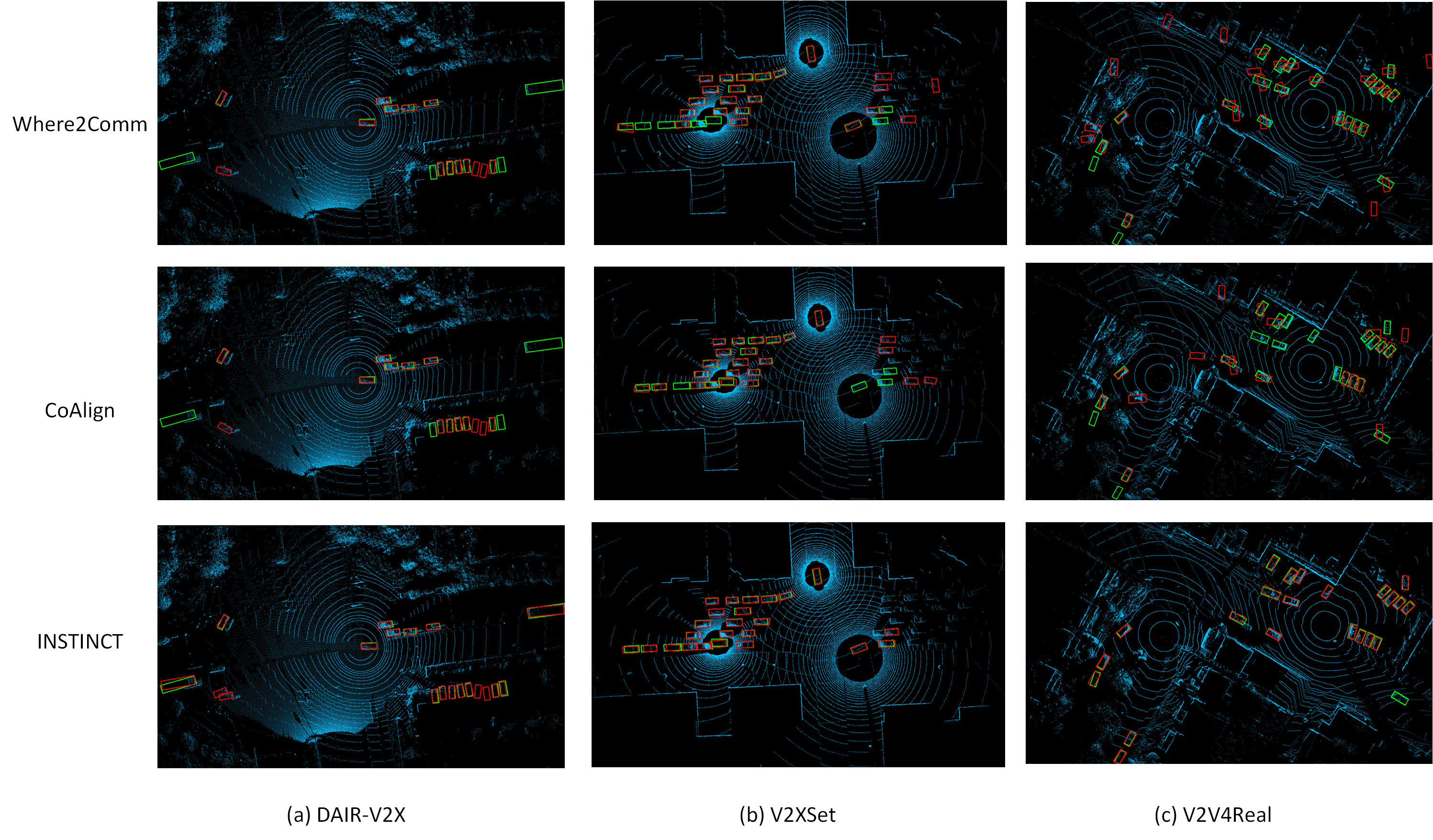}
    \caption{
    Visual comparison on different datasets.
    \textcolor{green}{Green} box denotes ground-truth, and \textcolor{red}{red} box denotes detection result.
    }
    \label{fig:enter-label}
\end{figure*}

\subsection{Adaptive Adjustment of GT Sampling}\label{GT_sampling}
While the proposed design enables effective local instance fusion, viewpoint diversity and scene sparsity often yield limited samples after dual-stage filtering and DDR, compromising model training. To resolve this, we develop Co-GT Sampling specifically designed for cooperative scenarios. In fact, DI-V2X \cite{li2024di} is the first work to apply this method in cooperative perception scenarios, but it uses it solely to bridge the domain gap between the teacher and student models, focusing primarily on domain adaptation in knowledge distillation.

Our implementation involves: 
1) Object-Centric Database: We built an object-centered point cloud database by point cloud cropping guided by ground-truth labels and cross-agent mixing.
2)
Ego-Centric Sampling:
Samples are drawn from the aforementioned database, then we validate them via spatial consistency verification to ensure their validity. 
Specifically, the sampled labels must neither overlap with each other nor with existing ground-truth annotations.
3) 
Scenario Synchronization:
To maintain scene consistency, we iterates other agents in the collaborative scenario, projects their point clouds into the ego vehicle's coordinate system, and determines whether any point cloud objects should be inserted. 
Eligible point clouds are then incorporated into the ego point cloud, and their label information is synchronized accordingly.
4)
Coordinate Restoration:
The processed agent point clouds and the augmented labels are inverse-transformed back to the agent's original coordinate system for single-agent detection supervision.

\subsection{Detection Head}
The obtained $\mathbf{Q}_i^{final}$ will be fed into the final output head in Step. \ref{final}. The parameters of this output head are independent, and it uses a separate FFN to perform predictions for classification and regression tasks. It is worth noting that we also employ the improved MAL loss for the classification task.


\section{Experiments}

\subsection{Setup}

{\bf Dataset.}
We conducted comprehensive experiments on 3 datasets to evaluate our proposed method, including DAIR-V2X \cite{yu2022dair}, V2XSet \cite{xu2022v2x} and V2V4Real \cite{xu2023v2v4real}.
\textbf{Dair-V2X} is a public real-world dataset for research on V2X autonomous driving.
It includes 71,254 LiDAR frames and 71,254 Camera frames.
Each sample scenario includes a vehicle, an infrastructure and its corresponding annotations.
\textbf{V2XSet} is a large-scale V2X perception dataset created by CARLA \cite{dosovitskiy2017carla} and OpenCDA \cite{xu2022opv2v}.
It is consists with 11,447 frames, including V2V cooperation and V2I cooperation.
\textbf{V2V4Real} is the first large-scale real-world dataset for vehicle-to-vehicle collaborative perception. Comprising 410 kilometers of driving coverage, V2V4Real captures multimodal sensor data including 20K LiDAR point clouds, 40K high-resolution RGB images, and 240K precisely annotated 3D bounding boxes, establishing a comprehensive benchmark for V2V cooperative perception research.
\\
{\bf Evaluation metrics.}
Following \cite{hu2022where2comm, chen2023transiff}, we use Average Precisions (AP) at Intersection-over-Union (IoU) thresholds of 0.5 and 0.7 to evaluate perception performance.
The communication bandwidth is counted in bytes and represented using a logarithm with base 2.
\\
{\bf Implementation details.}
All experiments with INSTINCT were conducted on one NVIDIA A100 GPU, using SECOND \cite{yan2018second} as the 3D backbone. 
The grid resolution is set to 0.1 m in length, width, and height (with the height adjusted to 0.2 m for V2V4Real).
We utilized ConQueR \cite{zhu2023conquer} to generate single-agent instances. 
For classification, we employed a combination of Focal Loss and MAL Loss, while L1 loss was used for regression. 
We use Adam optimizer \cite{kinga2015method} with an initial learning rate of 0.001, and the learning rate is adjusted following the OneCycle \cite{smith2019super} strategy. 
Finally, as training approached convergence, we use the Fade strategy \cite{wang2021pointaugmenting} to obtain a data distribution that better reflects real-world conditions.

\begin{figure}[htbp]
  \centering
  \includegraphics[width=\columnwidth, trim=46pt 25pt 56pt 35pt, clip]{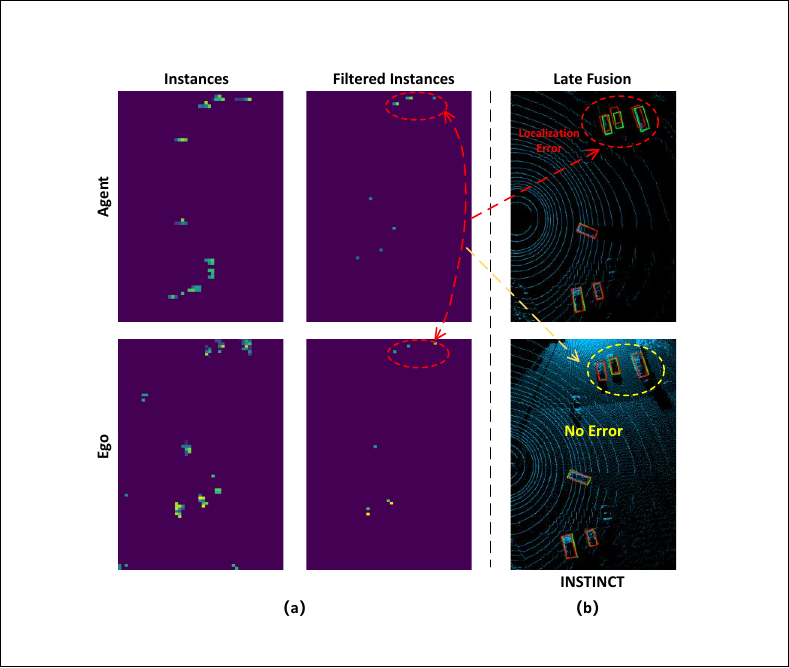}
  \caption{\textbf{Visualization of Late Fusion (ConQueR-based) vs. INSTINCT.}
(a) Instance feature filtering: agent vs. ego; (b) Detection results: Late Fusion vs. INSTINCT.}
  \label{fig:vis_feat}
\end{figure}

\subsection{Quantitative Results}
\textbf{Comparison of performance.}
To evaluate the performance of the INSTINCT model, we conduct multi-model comparative experiments on three benchmark datasets: DAIR-V2X, V2XSet, and V2V4Real. 
The experiments includes Three types of baselines: a No Fusion baseline without collaborative perception, a Late Fusion method that performs post-fusion based on detection results and six advanced intermediate fusion models including V2VNet, V2XViT, DiscoNet, Where2comm and CoAlign as comparative benchmarks. 
Detailed experimental data can be found in Tab. \ref{tab_performace}.
The results indicate that INSTINCT demonstrates significant performance advantages across all three datasets. 
On the real-world datasets DAIR-V2X and V2V4Real, the AP@0.7 metric surpasses the current best models by absolute margins of 13.23\% and 33.08\%, respectively.
And on the simulated dataset V2XSet, a 2.81\% performance improvement is maintained, albeit with a relatively limited gain. 
We attribute these differences to the following factors: 
(1)
Data characteristic disparities. 
As a simulated dataset, V2XSet exhibits relatively homogeneous scene distributions and object structures, enabling most comparative models to reach a high performance ceiling. 
However, it is noteworthy that INSTINCT also has limitations when detecting targets in overly sparse point clouds; 
(2)
Validation of architectural advantages. 
In contrast to conventional feature-level interaction methods, the instance-level interaction mechanism employed by INSTINCT shows stronger adaptability in complex real-world scenarios.
It is substantiated by the significant performance leaps observed on DAIR-V2X and V2V4Real.
For a more in-depth analysis of the model's performance, supplementary analysis is provided in the Appendix.
\\
\textbf{Comparison of Communication Bandwidth.}
Tab. \ref{tab_performace} also summarizes the communication bandwidth of each model.
The results indicate that while maintaining optimal performance, INSTINCT requires significantly lower communication bandwidth compared to the competing models. 
Specifically, 1) although the Late Fusion method achieves minimal communication overhead by transmitting low-dimensional bounding box data (averaging of 0.8 KB per frame), it relies on post-NMS fusion mechanism to prevent the calibration of overlapping detection results, leading to a notable performance degradation; 
2) in contrast, INSTINCT utilizes medium-dimensional instance feature transmission (averaging 16 KB per frame) and achieves detection calibration through high-dimensional feature fusion, thereby striking the best balance between communication efficiency and detection accuracy.
\\
\textbf{Robustness to Pose Errors.}
To evaluate the model's stability under realistic noise environments, we assessed its performance at varying pose noise intensities on the DAIR-V2X dataset (see Tab. \ref{tab_noise}). 
To simulate pose noise, Gaussian noise was added during inference to each agent's 2D position (with $\mu_t=0\text{ m}$ and $\sigma_t\in[0\text{ m},0.5\text{ m}]$) and yaw angle (with $\mu_r=0$ and $\sigma_r\in[0^{\circ},0.5^{\circ}]$), emulating real-world localization errors. 
The results shows that INSTINCT consistently maintains optimal detection accuracy across different noise levels, demonstrating exceptional environmental robustness.

\subsection{Qualitative Results}
\textbf{Detection Performance Visualization.} Fig. \ref{fig:enter-label} illustrates a comparison between the detection results of INSTINCT and the current state-of-the-art models, including Where2comm \cite{hu2022where2comm}, CoAlign \cite{lu2023robust} \etc. 
Visual analysis indicates that INSTINCT consistently achieves lower false positive rates across all three datasets. 
Moreover, as measured by IoU, the alignment between its predicted bounding boxes and GT is significantly superior to that of the competing model—especially in scenarios with complex occlusions.
It demonstrates enhanced object discrimination capability.\\
\textbf{Late Fusion vs. INSTINCT: A Visual Analysis.}
The exceptional performance of INSTINCT stems from its core component—the ConQueR instance generator. 
As an advanced 3D object detector, we conducted an in-depth comparison of the performance differences between ConQueR-based late fusion and INSTINCT by visualizing the instance distribution in feature maps. 
As shown in Fig. \ref{fig:vis_feat}, a clear comparison between subfigures (a) and (b) reveals that in long-range object detection, the ConQueR-based late fusion method exhibits significant localization deviations. 
In contrast, INSTINCT, leveraging its unique collaborative calibration mechanism, effectively integrates detection results from both the agent and ego vehicles, achieving precise localization alignment. Further details regarding this section are provided in the appendix.

\begin{table}[h]
\begin{minipage}{\columnwidth} 
\resizebox{\columnwidth}{!}{
\begin{tabular}{cccccc}
\hline
Noise Level                        & \multirow{2}{*}{0.0/0.0} & \multirow{2}{*}{0.1/0.1} & \multirow{2}{*}{0.2/0.2} & \multirow{2}{*}{0.3/0.3} & \multirow{2}{*}{0.4/0.4} \\
$\sigma_t/\sigma_r(m/\circ)$       &                          &                          &                          &                          &                          \\ \hline
\multicolumn{1}{c|}{Method/Metric} & \multicolumn{5}{c}{AP@0.5 ↑}                                                                  \\ \hline
\multicolumn{1}{c|}{V2VNet}              & 0.6572 & 0.6476 & 0.6336 & 0.6144 & 0.5974                       \\
\multicolumn{1}{c|}{V2XViT}              & 0.7349          & 0.7313          & 0.7190          & 0.7021          & 0.6914                       \\
\multicolumn{1}{c|}{DiscoNet}            & 0.7469          & 0.7436          & 0.7326          & 0.7205          & 0.7071                       \\
\multicolumn{1}{c|}{Where2comm}          & 0.7901          & 0.7847          & 0.7619          & 0.7301          & 0.7040                       \\
\multicolumn{1}{c|}{CoAlign}          & 0.7797          & 0.7758          & 0.7508          & 0.7220          & 0.6989                       \\
\multicolumn{1}{c|}{\textbf{Ours}}       & \textbf{0.8191} & \textbf{0.8160} & \textbf{0.7782} & \textbf{0.7327} & \textbf{0.7144}              \\ \hline
Method/Metric                            & \multicolumn{5}{c}{AP@0.7 ↑}                                                                    \\ \hline
\multicolumn{1}{c|}{V2VNet}              & 0.4982 & 0.4879 & 0.4705 & 0.4624 & 0.4512                       \\
\multicolumn{1}{c|}{V2XViT}              & 0.5675          & 0.5642          & 0.5550          & 0.5445          & 0.5363                       \\
\multicolumn{1}{c|}{DiscoNet}            & 0.5920          & 0.5891          & 0.5812          & 0.5744          & 0.5680                       \\
\multicolumn{1}{c|}{Where2comm}          & 0.6649          & 0.6403          & 0.6021          & 0.5718          & 0.5587                       \\
\multicolumn{1}{c|}{CoAlign}          & 0.6547          & 0.6327          & 0.5933          & 0.5727          & 0.5604                       \\
\multicolumn{1}{c|}{\textbf{Ours}}       & \textbf{0.7529} & \textbf{0.7238} & \textbf{0.6451} & \textbf{0.6105} & \textbf{0.5985}              \\ \hline
\end{tabular}}
\caption{
Performance on DAIR-V2X dataset with pose noise.
}
\label{tab_noise}
\end{minipage}
\end{table}
\subsection{Ablation Study}
To validate the effectiveness of each module, we conducted systematic ablation experiments on the DAIR-V2X dataset (see Tab. \ref{tab_ablation}).
The results are as follows:  
1) 
When all INSTINCT structures are removed, the model degrades to a ConQueR-based late fusion baseline, resulting in a significant performance drop.  
2) 
Enabling the quality-aware filtering module reduces the communication demand to 1/17 of the baseline (a reduction of approximately 94.1\%).
3) 
Introducing a DDR module to separate collaboration-relevant and irrelevant instances not only recovers the baseline performance while maintaining low bandwidth but also further increases the AP@0.7 by 3.43 percentage points compared to baseline.  
4) 
With the CDA module enabled, the interaction of mixed instances combined with DDR to eliminate interference boosts the AP@0.7 by 11.23 percentage points compared to baseline.  
5)
After adding the CALIF mechanism, the AP@0.7 further increases by 14.16 percentage points compared to baseline. 
The Co-GT Sampling strategy, by increasing the diversity of positive samples, drives an overall AP@0.7 improvement of 15.51 percentage points compared to baseline.
\begin{table}[ht]
\begin{minipage}{\columnwidth}
\resizebox{\columnwidth}{!}{
\begin{tabular}{ccccc|cc}
\hline
\multirow{2}{*}{QAF} & \multirow{2}{*}{DDR} & \multirow{2}{*}{CDA} & \multirow{2}{*}{GDA} & \multirow{2}{*}{GT} & \multirow{2}{*}{AP@0.5/0.7 ↑} & Comm   \\
                    &                     &                      &                     &                     &                             & (log2) \\ \hline
            &             &              &             & \multicolumn{1}{c|}{}           & 0.7302/0.5978        & 17.67 \\
\checkmark  &             &              &             & \multicolumn{1}{c|}{}           & 0.6962/0.6041        & 13.58 \\
\checkmark  & \checkmark  &              &             & \multicolumn{1}{c|}{}           & 0.7198/0.6321        & 13.58 \\
\checkmark  & \checkmark  & \checkmark   &             & \multicolumn{1}{c|}{}           & 0.7898/0.7101        & 13.58 \\
\checkmark  & \checkmark  & \checkmark   & \checkmark  & \multicolumn{1}{c|}{}           & 0.8109/0.7394        & 13.58 \\
\checkmark  & \checkmark  & \checkmark   & \checkmark  & \multicolumn{1}{c|}{\checkmark} & \textbf{0.8191}/\textbf{0.7529}  & 13.58\\ \hline
\end{tabular}
}
\caption{
Ablation study results of our proposed core components on DAIR-V2X dataset.
\textbf{QAF} : Quality-Aware Filtering in \ref{qulity filtering}; \textbf{DDR} : Dual-Branch Detection Routing in \ref{CDA}; \textbf{CDA} : Cross-Domain Adaption in \ref{instance_fusion}; \textbf{GDA} : Gaussian Distance Attention in \ref{instance_fusion}; \textbf{GT} : Co-GT Sampling in \ref{GT_sampling}.
}\label{tab_ablation}
\end{minipage}
\end{table}
\section{Conclusion}
We propose INSTINCT, an instance-level interaction architecture based on LiDAR-V2X Systems.
To ensure high-quality instances, we design a quality-aware filtering module. 
INSTINCT further employs a dual-branch detection strategy that decouple collaborative related and unrelated instances, thereby preventing interference. 
Moreover, during the fusion of hybrid instances, we introduce a cross-agent instance interaction mechanism that incorporates both domain gap compensation and a local interaction attention mechanism based on Gaussian distances. 
INSTINCT effectively integrates instances transmitted by multiple agents, and experimental results shows that it achieves state-of-the-art performance on benchmark datasets while maintaining an extremely low communication overhead.
\section*{Acknowledgements}
This work was supported in part by the National Natural Science Foundation of China (62072321), the Science and Technology Program of Jiangsu Province (BZ2024062), the Natural Science Foundation of the Jiangsu Higher Education Institutions of China (22KJA520007), Suzhou Planning Project of Science and Technology (2023ss03). 
{
    \small
    \bibliographystyle{ieeenat_fullname}
    \bibliography{main}
}
\clearpage
\setcounter{section}{0}
\setcounter{table}{0}
\setcounter{figure}{0}
\maketitlesupplementary


\section{More details}
\subsection{Implement details}
INSTINCT employs SECOND \cite{yan2018second} as the 3D backbone and uses ConQueR \cite{zhu2023conquer} as the single-agent baseline to provide instance-level features.
We set the number of layers in both the encoder and decoder within ConQueR to 3.
Moreover, in the final layer of the decoder, the Focal Loss used for classification supervision is replaced with MAL Loss \cite{huang2024deim}, with the corresponding $\gamma$ set to 1. 
In the subsequent filtering operation, we set the threshold to 0.1, which is identical to the $\lambda$ value used in our Dual-branch Detection Routing (DDR). 

After the collaborative output head, all instances are aggregated, and a unified detection head, which is composed of a 3-layer FFN, is used to generate the final output.
Notably, for classification supervision at this stage, we also employ MAL Loss configured the same way as in the decoder's final layer to penalize detection outputs that exhibit low IoU yet high confidence scores.

Both single-agent and collaborative outputs are supervised. 
As described in DETR \cite{carion2020end}, we utilize Hungarian matching for loss computation, and the number of queries is initialized to 300.

\textbf{Technical clarification of Co-GT Sampling strategy:} Co-GT Sampling constructs an object database $D = \{(P^{(i)}, L^{(i)})\}$, where $P^{(i)}$ and $L^{(i)}$ denote the $i$-th cropped object point cloud and its label. A subset $\mathcal{O} = \{o^{(j)}\} \subset D$ is selected such that $\mathrm{IoU}(o^{(j)}, o^{(k)}) = \mathrm{IoU}(o^{(j)}, \mathrm{GT}) = 0$ for all $j \neq k$, ensuring no overlap with other samples or existing GTs.
For each agent $a$, its point cloud $P_a$ and labels $L_a$ are transformed to the ego frame via $T_{e \leftarrow a}$, merged with $\mathcal{O}$ and corresponding labels, then inverse-transformed by $T_{a \leftarrow e}$ to produce the augmented pair $(P_a^{\mathrm{aug}}, L_a^{\mathrm{aug}})$:
$
(P_a, L_a) \xrightarrow{T_{e \leftarrow a}} (P_a \cup \mathcal{O},\, L_a \cup L_{\mathcal{O}}) \xrightarrow{T_{a \leftarrow e}} (P_a^{\text{aug}}, L_a^{\text{aug}}).
$
Legal samples $\mathcal{O}$ are thus aligned in the ego frame, shared across agents, and restored to local frames for conflict-free, consistent augmentation.

\subsection{Experiments details}
\textbf{Sufficient repetition of experiments (mean ± std):} All results in the manuscript are averaged over 5 runs with random seeds. 
We further reran 10 times on DAIR-V2X, yielding AP@0.5 = 0.8287 $\pm$ 0.0048 and AP@0.7 = 0.7539 $\pm$ 0.0067, confirming INSTINCT's consistent superiority over all baselines. 

\textbf{Runtime and Bandwidth Details:}
Excluding the 3D backbone, collaborative modules incur inference time of 110/120/62\,ms (INSTINCT/V2X-ViT/Where2comm) on an A100, with total time of 175/110/77\,ms. This indicates that INSTINCT’s backbone is the primary speed bottleneck, and its collaborative modules remain inefficient. 
Bandwidth averages 16\,KB/frame across datasets; on DAIR-V2X it is 12\,KB (median 11, range 0–36, variance 8), scaling linearly with object count, while other methods grow quadratically with sensing range.

\textbf{Comparison to TransIFF \cite{chen2023transiff} \& QUEST \cite{fan2024quest} \& ACCO \cite{yang2025discretization}:}
Although TransIFF achieves low bandwidth, it falls short of mainstream SOTA performance and is limited to vehicle-infrastructure collaboration. 
Our reproduction using the same single detector as INSTINCT yields AP@0.5/0.7 of 0.6919/0.5879 on DAIR-V2X (higher than reported), with a mean bandwidth of 13.58, same as INSTINCT due to identical detector use. 
Thus, aside from bandwidth, TransIFF underperforms compared to SOTA methods.
QUEST and ACCO also adopt instance-level interaction, but are designed for camera-based perception, whereas INSTINCT focuses on LiDAR. 
Although INSTINCT outperforms them, direct comparison is unfair due to the modality gap.

\textbf{Performance gain: collaboration or single detector?}
We addressed this in Appendix Sec.\ref{compare_3d}, comparing against two baselines: late fusion with the best single detector, and INSTINCT using the same weights. 
INSTINCT outperforms late fusion by 12.17\% at AP@0.5 and 25.95\% at AP@0.7, demonstrating its effectiveness in calibrating detections.
We replace the single detector with Voxel-DETR (whose mAP@0.7 is $\sim$5\% lower than ConQueR) to test INSTINCT on DAIR-V2X. The AP@0.7 drops slightly to 0.7303 (by 2\%), while the communication cost increases to 13.69. 
It indicates that INSTINCT is robust to weaker detectors, though it needs higher communication overhead.
\section{Comparison with other 3D Detectors}\label{compare_3d}
Advanced single-vehicle models provide sufficiently accurate instance features for INSTINCT.
To demonstrate INSTINCT’s ability to effectively fuse these instance features, we compared its performance with other state-of-the-art 3D detectors that employ direct late fusion. 
Additionally, we evaluated the impact of using MAL loss versus quality prediction on detection performance. 
In the quality prediction approach, an extra 3-layer FFN is used to predict the IoU, and the IoU between positively matched samples and the ground truth (obtained through Hungarian matching) serves as the supervision signal.

\begin{table*}[ht]
\centering
\begin{tabular}{ccc|cc}
\hline
                          &                                 &                          & DAIR-V2X                             & V2XSet                 \\ \cline{4-5} 
\multirow{-2}{*}{Model}   & \multirow{-2}{*}{Fusion Method} & \multirow{-2}{*}{Quality Aware Method} & \multicolumn{2}{c}{AP@0.5/0.7}                                \\ \hline
                          &                                 & \textbackslash{}         & {\color[HTML]{262626} 0.7252/0.5840} & 0.9310/0.8974          \\
                          &                                 & MAL                      & {\color[HTML]{262626} 0.7302/0.5978} & 0.9489/0.9085          \\
\multirow{-3}{*}{ConQueR} & \multirow{-3}{*}{Late Fusion}   & MAL + Quality Prediction & {\color[HTML]{262626} 0.7213/0.5717} & \textbf{0.9593/0.9242} \\ \hline
SEED                      & Late Fusion                     & \textbackslash{}         & {\color[HTML]{262626} 0.7332/0.5961} & 0.9377/0.9064          \\ \hline
                          &                                 & \textbackslash{}         & 0.8077/0.7415                        & 0.9123/0.8688          \\
                          &                                 & MAL                      & \textbf{0.8191/0.7529}               & 0.9229/0.8731          \\
\multirow{-3}{*}{INSTINCT} & \multirow{-3}{*}{Instance}     & MAL + Quality Prediction & 0.8111/0.7401                        & 0.9272/0.8822          \\ \hline
\end{tabular}
\caption{\textbf{Comparison with other 3D Detectors.}}\label{tab:compare}
\end{table*}
Based on the results in Tab. \ref{tab:compare}, the performance of the INSTINCT model across different datasets exhibits the following characteristics:
\begin{enumerate}
    \item \textbf{Performance Comparison on the DAIR-V2X Dataset.}
    INSTINCT’s late fusion performance is significantly superior to that of ConQueR and SEED . 
    This demonstrates that under the complex data distributions encountered in real-world scenarios (e.g., variations in illumination, occlusions), the instance fusing mechanism in INSTINCT can more effectively integrate information from multiple agents. 
    In contrast, traditional late fusion methods lack effective calibration capabilities, struggle to handle the noise interference present in real environments.
    \item \textbf{Analysis of the V2XSet Simulation Dataset.}
    Although INSTINCT achieves the best performance in the intermediate fusion model, its performance is slightly lower than the late fusion results of ConQueR and SEED \cite{liu2024seed} in V2XSet. We attribute this to two factors:  
    \begin{itemize}
        \item \textbf{Simplicity of the Data Distribution.}
        As a simulated dataset, the distribution of 3D detection targets in V2XSet is highly idealized, resulting in near-saturation of prediction accuracy among the agents. 
        In such cases, maintaining the original predictions can actually be the optimal strategy. 
        \item \textbf{Model Training Challenges.}
        INSTINCT is designed to learn “how to fuse” instance features through training, but the strong consistency of “unchanged” targets in simulated data makes it difficult for the model to converge to the desired state.
        \item \textbf{Practical Implication.}
        Since collaborative perception is ultimately applied in real autonomous driving scenarios (e.g., the complex road tests in DAIR-V2X), the advantages of INSTINCT on real-world data hold greater engineering value.
    \end{itemize}
    \item \textbf{Trade-offs Between the Loss Function and the Quality Prediction Module.}
    \begin{itemize}
        \item \textbf{MAL Loss.}
        This loss function consistently boosts performance in both the multi-agent baseline models and INSTINCT, validating its effectiveness in enhancing instance selection through adversarial learning. 
        \item \textbf{Quality Prediction (IoU).}
        On the V2XSet dataset, the simplicity of the targets results in an 1.5\% gain. 
        However, on the DAIR-V2X dataset, the limited data scale leads to prediction biases that cause a performance drop. 
        In the current approach, we rely solely on MAL Loss for quality-aware filtering. 
        Future work will focus on improving the robustness of IoU prediction through a pre-training and fine-tuning strategy or uncertainty modeling.
    \end{itemize}
\end{enumerate}
This analysis highlights the strengths of INSTINCT, particularly in real-world scenarios, and outlines potential avenues for further enhancing the model's robustness and performance.

\section{Effectiveness of CDA and GDA}
\begin{figure*}[h]
  \centering
  \includegraphics[width=\textwidth]{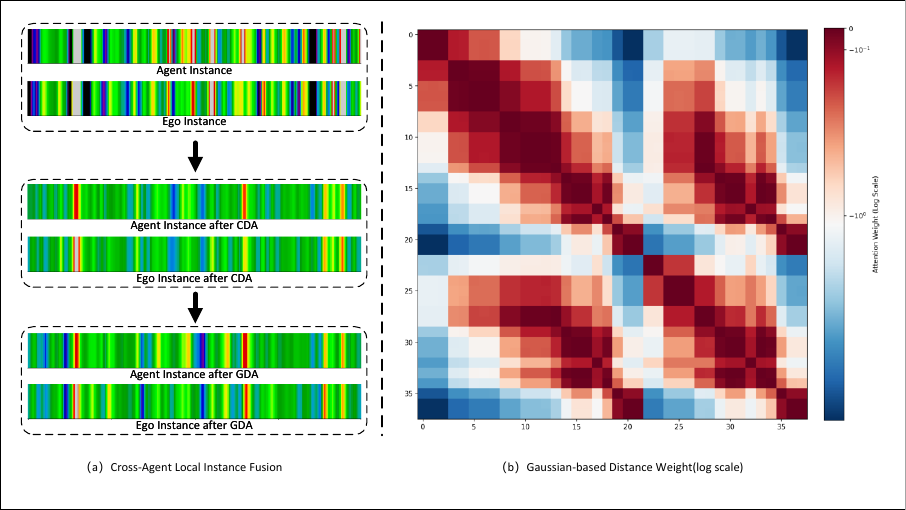}
   \caption{\textbf{Visualization of effectiveness of CDA and GDA.}
   (a) shows the the query after CDA and GDA.
   (b) shows the relationship between Gaussian distance and weight.
   }
   \label{fig:eff}
\end{figure*}
To more intuitively illustrate the roles of the two modules, CDA and GDA, in Cross-Agent Local Instance Fusion (CALIF), we provide a visualization of the fused instances from the ego and agents.
As shown in Fig. \ref{fig:eff}(a), after applying CDA, the numerical distributions of the ego and  agent instances become more similar. 
Following GDA, both instances undergo self-calibration based on each other’s information.
Notably, ego instance features clearly acquire additional information from the agent instance features. 
In Fig. \ref{fig:eff}(b), we visualize the weights corresponding to GDA, which indicate the degree of mutual attention between the mixed instances. 
The weight matrix generally appears symmetric, suggesting a tight correspondence between the instance features. 
It is important to note that similar colors do not imply identical values, but rather comparable attention levels. 
The attention weights provided by GDA facilitate effective local fusion of the fused instance features.
\section{More Visualization Results}

\begin{figure*}[h]
  \centering
  \includegraphics[width=\textwidth]{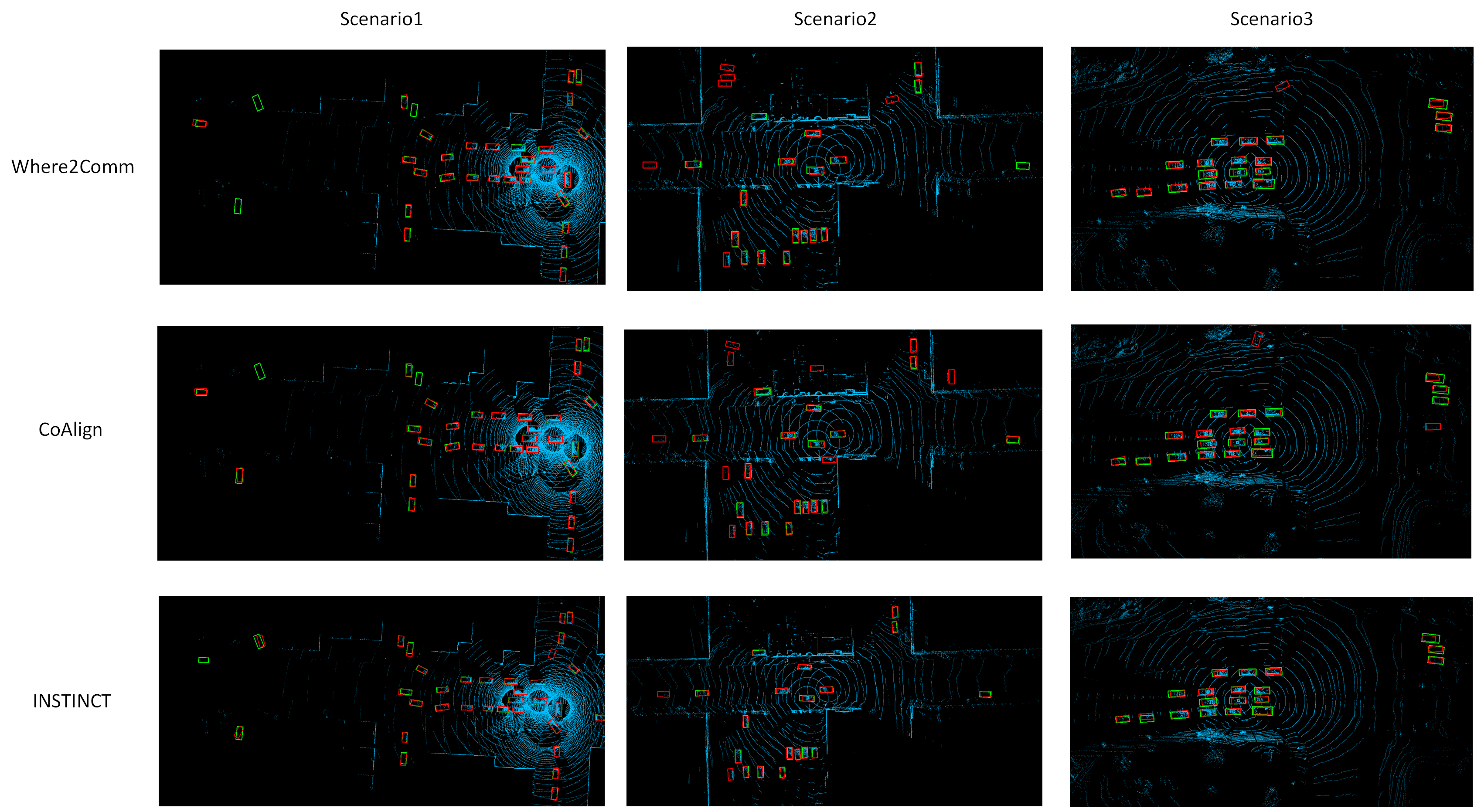}
   \caption{\textbf{Comparison of BEV Visualization of Different Models in Different Scenarios.}}
   \label{fig:vis2d}
\end{figure*}

\begin{figure*}
    \centering
    \includegraphics[width=\textwidth]{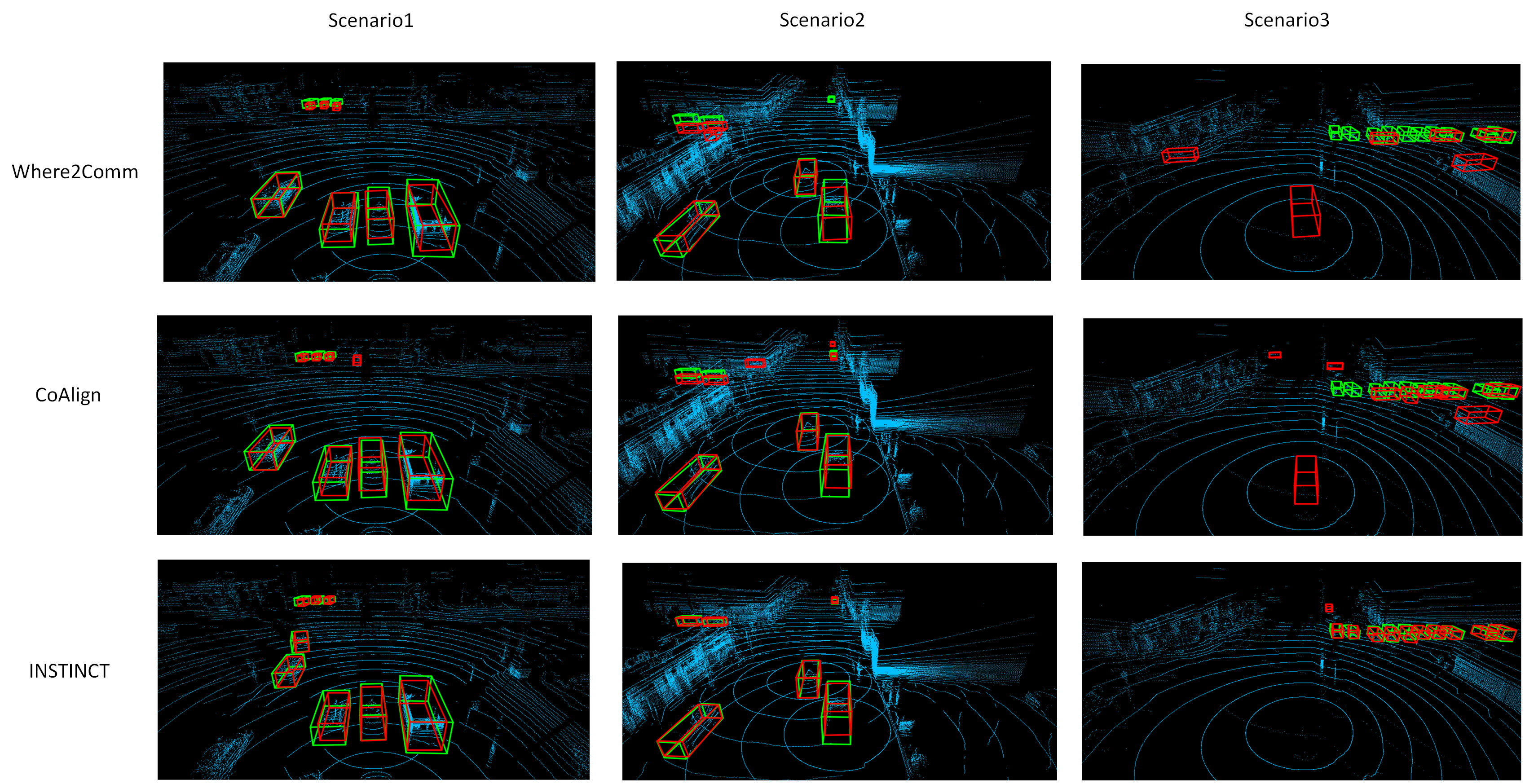}
    \caption{\textbf{Comparison of 3D Visualization of Different Models in Different Scenarios.}}
    \label{fig:vis3d}
\end{figure*}

We provide additional visual comparisons of INSTINCT on V2XSet and V2V4Real in the Fig. \ref{fig:vis2d} and Fig. \ref{fig:vis3d}.
In addition to BEV detection results, we also present visualizations of the 3D detection result. 
All visualizations indicate that INSTINCT exhibits stronger calibration capability and more stable performance compared to state-of-the-art intermediate fusion methods.


\end{document}